\documentclass[sn-basic]{sn-jnl}

\usepackage[T1]{fontenc}
\usepackage[retain-explicit-plus]{siunitx}
\usepackage{xcolor}
\usepackage{booktabs}
\usepackage{tabu}
\usepackage{tabularx}
\usepackage{tabulary}
\usepackage{multirow,bigdelim}
\usepackage{rotating}
\usepackage{makecell}
\usepackage{threeparttable}
\usepackage{listings}
\usepackage{enumitem}
\usepackage{comment}
\usepackage{tablefootnote}
\usepackage{subcaption}
\usepackage{enumitem,amssymb}
\usepackage{graphicx}
\usepackage{makecell}
\usepackage{lineno}

\usepackage{blindtext}
\usepackage{hyperref}
\usepackage{nameref}

\newcounter{mylabelcounter}

\makeatletter
\newcommand{\labelText}[2]{%
\refstepcounter{mylabelcounter}%
\immediate\write\@auxout{%
 \string\newlabel{#2}{{\unexpanded{#1}}{\thepage}{{\unexpanded{#1}}}{mylabelcounter.\number\value{mylabelcounter}}{}}%
}%
}
\makeatother

\newlist{todolist}{itemize}{2}
\setlist[todolist]{label=$\square$}

\jyear{2022}

\begin{document}

This version of the article has been accepted for publication, after peer review but is not the Version of Record and does not reflect post-acceptance improvements, or any corrections. The Version of Record is available online at: https://doi.org/10.1007/s10994-024-06527-w.

\title[Exposing and Explaining Fake News On-the-fly]{Exposing and Explaining Fake News On-the-fly}

\author*[1]{\fnm{Francisco} \sur{de Arriba-Pérez}}\email{farriba@gti.uvigo.es}
\equalcont{These authors contributed equally to this work.}

\author[1]{\fnm{Silvia} \sur{García-Méndez}}\email{sgarcia@gti.uvigo.es}
\equalcont{These authors contributed equally to this work.}

\author[2]{\fnm{Fátima} \sur{Leal}}\email{fatimal@upt.pt}
\equalcont{These authors contributed equally to this work.}

\author[3,4]{\fnm{Benedita} \sur{Malheiro}}\email{mbm@isep.ipp.pt}
\equalcont{These authors contributed equally to this work.}

\author[1]{\fnm{Juan Carlos} \sur{Burguillo}}\email{J.C.Burguillo@uvigo.es}
\equalcont{These authors contributed equally to this work.}

\affil[1]{atlanTTic, University of Vigo, Information Technologies Group, Campus Universitario de Vigo, Lagoas-Marcosende, 36310, Vigo, Spain}

\affil[2]{REMIT, Universidade Portucalense, Rua Dr. António Bernardino de Almeida, 4200-072 Porto, Portugal}

\affil[3]{ISEP, Polytechnic of Porto, Rua Dr. António Bernardino de Almeida, 4249-015 Porto, Portugal}

\affil[4]{INESC TEC, Campus da Faculdade de Engenharia da Universidade do Porto, Rua Dr. Roberto Frias, 4200-465 Porto, Portugal}

\abstract{Social media platforms enable the rapid dissemination and consumption of information. However, users instantly consume such content regardless of the reliability of the shared data. Consequently, the latter crowdsourcing model is exposed to manipulation. This work contributes with an explainable and online classification method to recognize fake news in real-time. The proposed method combines both unsupervised and supervised Machine Learning approaches with online created lexica. The profiling is built using creator-, content- and context-based features using Natural Language Processing techniques. The explainable classification mechanism displays in a dashboard the features selected for classification and the prediction confidence. The performance of the proposed solution has been validated with real data sets from Twitter and the results attain \SI{80}{\percent} accuracy and macro \textit{F}-measure. This proposal is the first to jointly provide data stream processing, profiling, classification and explainability. Ultimately, the proposed early detection, isolation and explanation of fake news contribute to increase the quality and trustworthiness of social media contents.}


\maketitle


\section{Introduction}

In social media, information is shared collaboratively through platforms like Facebook\footnote{Available at \url{https://www.facebook.com}, June 2023.}, Twitter\footnote{Available at \url{https://twitter.com}, June 2023.}, or Wikinews\footnote{Available at \url{https://www.wikinews.org}, June 2023.}. Such platforms enable the rapid dissemination of information regardless of its trustworthiness, leading to instant consumption of non-curated news. The negative consequence of this openness of social media platforms is the spread of false information disguised as truth, \textit{i.e.}, fake news. Fake news can be defined as deceptive posts with an intention to mislead consumers in their purchase or approaching the context of misinformation and disinformation \citep{Xiao2020}. Specifically, while misinformation is an inadvertent action, disinformation is a deliberate creation/sharing of false information. The authenticity and intention can be distinguished as: (\textit{i}) non-factual and mislead, \textit{i.e.}, deceptive news and disinformation; (\textit{ii}) factual and mislead (cherry-picking); (\textit{iii}) undefined and mislead (click-bait); and (\textit{iv}) non-factual and undefined, \textit{i.e.}, misinformation.

Misinformation and fake news are characterized by their big volume, uncertainty, and short-lived nature. Furthermore, they disseminate faster and further on social media sites causing serious impact on politics and economics \citep{tandoc2019facts}. Accordingly, the report on digital transformation of media and the rise of disinformation/fake news of the European Union (EU) \citep{martens2018digital} reinforces the need to strengthen trust in digital media.

This work contributes with a real-time explainable classification method to recognize fake news, promoting trust in digital media as suggested by the SocialTruth project\footnote{Available at \url{http://www.socialtruth.eu/index.php/documentation}, June 2023.}. In fact, the early discarding of fake news has a positive impact on both information quality and reliability. The proposed method employs stream processing, updating the profiling and classification models on each incoming event. The profiling is built using side-based (related to the creator user and propagation context) and content-based features (extracted from the news text through Natural Language Processing (\textsc{nlp}) techniques), together with unsupervised methods, to create clusters of representative features. The classification relies on stream Machine Learning (\textsc{ml}) algorithms to classify in real-time the nature of each cluster. Finally, the proposed method includes an explanation mechanism to detail why an event has been classified as fake or non-fake. The explanations are presented visually and in natural language on the user dashboard.

The rest of this paper is organized as follows. Section~\ref{sec:2} overviews the relevant work on fake news concerning the profiling, classification and detection tasks. Section~\ref{sec:proposed_method} introduces the proposed method, detailing the data processing and stream-based classification procedures along with the online explainability. Section~\ref{sec:experimental_results} describes the experimental set-up and the empirical evaluation results considering the online classification and explanation. Finally, Section~\ref{sec:conclusion} concludes and highlights the achievements and future work.

\section{Related work}
\label{sec:2}

Social media plays a crucial role in news consumption due to its low cost, easy access, variety, and rapid dissemination \citep{Hu2014IEEE}. Indeed, social media is becoming an increasing source of breaking news. However, the fake news problem indicates that social platforms suffer from lack of transparency, reliability, and real-time modeling. In this context, fake news (misinformation/disinformation, such as rumor, deception, hoaxes, spam opinion, click-bait and cherry-picking) are false information created with the dishonest intention to mislead consumers \citep{Xiao2020,Choras:2021}. To characterize the nature of fake news and understand whether they result from inadvertent or deliberate action, it is necessary to establish their authenticity and the intention of the creator \citep{Shu2017}. In addition, social media streams are subject to feature variation over time \citep{Bondielli2019,Choras:2021}. Thus, the accurate detection of fake news in real time requires proper profiling and classification techniques. However, according to \cite{Shu:2022}, the current detection techniques are based on opaque models, leaving users clueless about classification outcomes. Consequently, the current work addresses transparency through explanations, reliability through fake news detection, and real-time modeling through incremental content profiling.

The following discussion compares existing works in terms of: (\textit{i}) stream-based profile modeling for fake detection; (\textit{ii}) stream-based classification mechanisms; and (\textit{iii}) transparency and credibility in detection tasks.

\subsection{Profiling}

Profiling methods model the stakeholders according to their contributions and interactions. Due to information sparsity, it is frequent to represent profiles using side and content information. In addition, in stream-based modeling, profiles are continuously updated and refined. To model fake news stakeholders, the literature contemplates multiple types of profiling methods: (\textit{i}) creator-based; (\textit{ii}) content-based; and (\textit{iii}) context-based. 

\begin{description}

\item [Creator-based] profiling focuses on both demographic and behavioral characteristics of the creator. Specifically, the literature contemplates account name, anomaly score\footnote{It is computed by the number of the user’s interaction in a time window divided by the user’s monthly average.}, credibility score, geolocation information, ratio between friends and followers, total number of tweets/posts, \textit{etc.} \citep{Castillo:2011,Zubiaga2017,Goindani2019,Vicario2019,Liu2020,Jang2021,Li2021,Silva:2021,Jain2022,Mosallanezhad2022}.

\item [Content-based] profiling explores textual features extracted from the post aiming to identify the meaning of the content. It can be obtained using linguistic and semantic knowledge, or style analysis via \textsc{nlp} approaches together with fact-checking resources\footnote{\textit{E.g.}, Classify.news, FackCheck.org, Factmata.com, Hoaxy.iuni.iu.edu, Hoax-Slayer.com, PolitiFact.com, Snopes.com, TruthOrFiction.com.}. Most of the revised works exploit this type of features. Therefore, content-based profiling encompasses:
\begin{itemize}
 \item \textit{Lexical and syntactical features} are properties related to the syntax, \textit{e.g.}, sentence-level features, such as bag-of-words approaches, $n$-grams, and part-of-speech. These features are exploited by \cite{Dong2020, Zhou2020Theory}. In addition, \cite{Vicario2019,Jang2021} compute the overall sentiment score of sentences.

 \item \textit{Stylistic features} provide emphasis and clarity to the text. Tweet-writing styles can be determined through: (\textit{i}) physical style analysis (\textit{e.g.}, number of adjectives, nouns, hashtags and mentions as well as emotion words and casual words); and (\textit{ii}) non-physical style analysis (\textit{e.g.}, complexity and readability of the text). The work by \cite{Jang2021} is a representative example of the physical style analysis.

 \item \textit{Visual features} describe the properties of images or videos used to ascertain the credibility of multimedia content. Visual features can: (\textit{i}) be purely statistic (\textit{e.g.}, number of images/videos); (\textit{ii}) represent distribution patterns; or (\textit{iii}) describe user accounts (\textit{e.g.}, background images). While \cite{Jang2021} compute statistic visual features, \cite{Liu2020} consider information from the user account. \cite{Li2021} verify if the image has been tampered, integrating this information as visual content, and \cite{Ying2021} combine textual with visual content to generate multi-level semantic features.
 \end{itemize}
 
\item [Context-based] profiling analyses both the surrounding environment and the creator engagements around the piece of information posted \citep{Castillo:2011,Goindani2019,Shu2019Beyond,Liu2020,Zhao2020,Jang2021,Li2021,Puraivan2021,Silva:2021,Song2021,Jain2022}. Specifically, it applies user-network analysis and distribution pattern analysis to obtain:

\begin{itemize}
 \item \textit{Network-based features} which aggregate similar online users in terms of location, education background, and habits \citep{Shu2019Beyond,Liu2020,Silva:2021}.
 
 \item \textit{Propagation-based features} that describe the dissemination of fake news based on the propagation graph as in the work by \cite{Mosallanezhad2022}. These may include, for an online account, the root degree, sub-trees number, the maximum/average degree and depth tree depth \citep{Castillo:2011,Jang2021} or the number of retweets/re-posts for the original tweet/post, the fraction of tweets/posts retweeted \citep{Zhao2020,Li2021}.
 
 \item \textit{Temporal-based features} which detail how two posts/tweets relate in time. They may comprise the posting frequency, the day of the week of the post \citep{Jang2021,Silva:2021}, the interval between two posts or even a complete temporal graph \citep{Song2021}.
\end{itemize}
\end{description}

\subsection{Classification}

Fake news detection is a classification task. The main news classification techniques in the literature encompass supervised, semi-supervised, unsupervised, deep learning, and reinforcement learning approaches. Deep learning, depending on the problem, can fall into the supervised or unsupervised classification scope \citep{Mathew2021}. Moreover, its high computational cost requires more computational resources than the corresponding traditional approaches, motivating a separate discussion.

\begin{description}

\item \textit{Supervised} classification is a widely used technique to map objects to classes based on numeric features or inputs (see Table \ref{tab:comparison}). The most frequently used supervised fake news detectors are Bayes, Probabilistic, Neighbor-based, Decision Trees, and Ensemble classifiers.

\item \textit{Semi-supervised} classification algorithms learn from both labeled and unlabeled samples. They are employed when it is difficult to annotate manually or automatically the samples. The works by \cite{Shu2019Beyond} and \cite{Dong2020} use supervised learning for fake news detection. 

\item \textit{Unsupervised} classification techniques group statistically similar unlabeled data based on underlying hidden features, using clustering algorithms or neural network approaches. The most commonly used cluster algorithms include \textit{k}-means, Iterative Self-Organizing Data Analysis Technique, and Agglomerative Hierarchical. \cite{Li2021,Puraivan2021} are representative examples of this approach.

\item \textit{Deep Learning} classification relies essentially on neural networks with three or more layers. In terms of fake news, deep learning has been employed mainly for text classification using Convolutional Neural Networks (\textsc{cnn}), Long Short Term Memory (\textsc{lstm}), and Recurrent Neural Networks (\textsc{rnn}) as in the works by \cite{Akinyemi2020,Nasir2021}.

\item \textit{Reinforcement Learning} classification works with unlabeled data \citep{sutton2018reinforcement}, but tends to be slow when applied to real-world classification problems \citep{dulac2021challenges}. While \cite{Goindani2019,Wang2020,Mosallanezhad2022} perform fake news detection through reinforcement learning, the most used technique is the Multivariate Hawkes Process (\textsc{mhp}) by \cite{Goindani2019}.

\end{description}

Classification can be performed offline or online. Offline or batch processes build static models from pre-existing data sets, whereas online or stream-based processes compute incremental models from live data streams in real-time.

\begin{description}

\item [Offline] classification divides the data set into training -- used to create the model -- and testing -- to assess the quality of the model -- partitions. The model remains static throughout the testing stage. This is the most popular fake news detection approach found in the literature.

\item [Online] classification mines data streams in real-time. Fake news, being dynamic sequences of data originated from multiple sources, \textit{i.e.}, the crowd, demand real-time processing. Typically, whenever new data arrive, the models are incrementally updated, enabling the generation of up-to-date classifications. To the best of the authors' knowledge, only \cite{Ksieniewicz2020} perform online fake news detection, processing samples as a data stream and considering concept drifts, \textit{i.e.}, that sample classification may naturally change over time.
\end{description}

Classification models can be interpretable and opaque. While opaque models behave as black boxes (\textit{e.g.}, standalone deep neural networks), interpretable models are self-explainable (\textit{e.g.}, trees- or neighbor-based algorithms). Interpretable classifiers explain classification outcomes \citep{Skrlj2021}, clarifying why a given content is false or misleading. More in detail, the explainable fake news detection framework by \cite{Shu:2019} integrates a news content encoder, a user comment encoder, and a sentence-comment co-attention network. The latter captures the correlation between news contents and comments and chooses the top-$k$ sentences and comments to explain the classification outcome. \cite{Zhou2020Theory} explore lexicon-, syntax-, semantic-, and discourse-level features to enhance the interpretablity of the models. \cite{Mahajan:2021} and \cite{Kozik:2022} adopt model agnostic interpretability techniques, such as Local Interpretable Model-agnostic Explanations (\textsc{lime}) \citep{Ribeiro:2016} and the Shapley Additive Explanations (\textsc{shap}) \citep{Lundberg:2017}, respectively. Finally, \cite{Silva:2021} provide explanations based on feature weights assigned to tweet/retweet nodes in the propagation patterns.

Table \ref{tab:comparison} provides an overview of the above works considering profiling (creator-, content-, and context-based), classification (supervised, semi-supervised, unsupervised, and reinforcement learning), processing (offline and online) and explainability. Summing up, this literature review shows that existing explainable fake news detectors explore creator-, content-, and context-based profiles, essentially adopt supervised classification and mostly implement offline processing.

\begin{table}[!htbp]
\centering
\footnotesize
\caption{Comparison of fake news detection approaches considering: (\textit{i}) profiling (creator, content, context), (\textit{ii}) classification (supervised, semi-supervised, unsupervised, reinforcement learning), (\textit{iii}) execution (offline, online), and (\textit{iv}) explainability (Ex.).}
\label{tab:comparison}
\begin{tabular}{lcccccccc}
\toprule
\textbf{Proposal} & \textbf{Profiling} & \textbf{Classification} & \textbf{Execution} & \textbf{Ex.}\\
\midrule

\cite{Castillo:2011} & \multirow{4}{*}{\shortstack{Creator\\ Content\\ Context}} & \multirow{4}{*}{Supervised} & \multirow{4}{*}{Offline} & \multirow{4}{*}{No}\\
\cite{Liu2020}\\
\cite{Jang2021}\\
\cite{Jain2022}\\
\midrule

\cite{Zubiaga2017} & Creator & \multirow{2}{*}{Supervised} & \multirow{2}{*}{Offline} & \multirow{2}{*}{No}\\
\cite{Vicario2019} & Content \\
\midrule

\multirow{2}{*}{\cite{Song2021}} & Content & \multirow{2}{*}{Supervised} & \multirow{2}{*}{Offline} & \multirow{2}{*}{No}\\
& Context\\
 \midrule

\cite{Akinyemi2020} & \multirow{5}{*}{Content} & \multirow{5}{*}{Supervised} & \multirow{5}{*}{Offline} & \multirow{5}{*}{No} \\
\cite{Silva2020} \\
\cite{Nasir2021} \\
\cite{Ying2021} \\
\cite{Galli2022} \\
\midrule

\cite{Zhao2020} & Context & Supervised & Offline & No\\
 \midrule

\cite{Dong2020} & Content & Semi-supervised & Offline & No\\
\midrule

\cite{Shu2019Beyond} & Context & Semi-supervised & Offline & No\\
\midrule

\multirow{2}{*}{\cite{Puraivan2021}} & Content & Unsupervised & \multirow{2}{*}{Offline} & \multirow{2}{*}{No}\\
& Context & \& supervised\\
\midrule

\multirow{3}{*}{\cite{Li2021}} & Creator & \multirow{3}{*}{Unsupervised} & \multirow{3}{*}{Offline} & \multirow{3}{*}{No}\\
 & Content \\
 & Context \\
\midrule

\multirow{2}{*}{\cite{Mosallanezhad2022}} & Creator & \multirow{2}{*}{\makecell{Reinforcement \\Learning}} & \multirow{2}{*}{Offline} & \multirow{2}{*}{No}\\
& Content\\\midrule

\multirow{2}{*}{\cite{Goindani2019}} & Creator & \multirow{2}{*}{\makecell{Reinforcement \\ Learning}} & \multirow{2}{*}{Offline} & \multirow{2}{*}{No}\\
& Context\\\midrule

\multirow{2}{*}{\cite{Wang2020}} & \multirow{2}{*}{Content} & \multirow{2}{*}{\makecell{Reinforcement \\ Learning}} & \multirow{2}{*}{Offline} & \multirow{2}{*}{No}\\
\\\midrule

\multirow{3}{*}{\cite{Silva:2021}} & Creator & \multirow{3}{*}{Supervised} & \multirow{3}{*}{Offline} & \multirow{3}{*}{Yes}\\
& Content \\
& Context \\
\midrule

\cite{Shu:2019} & \multirow{4}{*}{Content} & \multirow{4}{*}{Supervised} & \multirow{4}{*}{Offline} & \multirow{4}{*}{Yes}\\
\cite{Zhou2020Theory} \\
\cite{Mahajan:2021} \\
\cite{Kozik:2022} \\
\midrule

\cite{Ksieniewicz2020} & \multirow{1}{*}{Content} & \multirow{1}{*}{Supervised} & \multirow{1}{*}{Online} & \multirow{1}{*}{No}\\
\midrule

\multirow{3}{*}{\bf Current} & Creator & \multirow{3}{*}{\makecell{Unsupervised \\ \& Supervised}} & \multirow{3}{*}{Online} & \multirow{3}{*}{Yes}\\
& Content \\
& Context \\
 
\bottomrule
\end{tabular}
\end{table}

The most closely related works from the literature, considering the \textsc{pheme} experimental data used for design and evaluation, are the fake news classification solutions proposed by \cite{Zubiaga2017}, \cite{Akinyemi2020}, \cite{Ying2021}, and \cite{Jain2022}. Firstly, \cite{Zubiaga2017} experimented with sequential (Conditional Random Fields, Maximum Entropy and Enquiry-based) and non-sequential (Naive Bayes, Support Vector Machines (\textsc{svm}) and Random Forests (\textsc{rf})) classifiers. Secondly, \cite{Akinyemi2020} applied a \textsc{rf} model as the meta classifier trained with a stack-ensemble of \textsc{svm}, \textsc{rf}, and \textsc{rnn} models as base learners. Thirdly, \cite{Ying2021} presented a Multi-level Multi-modal Cross-attention Network for batch fake detection. Furthermore, \cite{Jain2022} employed a Hierarchical Attention Network (\textsc{han}) and a Multi-Layer Perceptron (\textsc{mlp}) trained with creator-, content-, and context-based features. The final prediction (fake or non-fake) combines both classifier outputs through a logical \textsc{or}. Nonetheless, all these solutions work offline without explaining the outcomes. In contrast, our work exploits a wide variety of profiling features (creator, content, and context), operates online and is able to explain the classification outcomes.

Similarly to our research, \cite{Puraivan2021} combineed both unsupervised and supervised techniques, for feature extraction (Principal Component Analysis and t-Distributed Stochastic Neighbor Embedding) and classification (optimized distributed gradient boosting), respectively. However, this offline work disregards the textual content of the news and lacks transparency.

Finally, the sole online system found explores fake news detection with Gaussian Naive Bayes, \textsc{mlp}, and Hoeffding Tree base learners independently and in ensembles \citep{Ksieniewicz2020}. Unfortunately, this work uses another data set collected by the authors and automatically labeled by BS Detector Chrome Extension. Profiles are exclusively based on content features and the outcomes are not explained.

\subsection{Research contribution}

As previously stated, this work contributes with an explainable classification method to recognize in real-time fake news and, thus, promote trust in digital media. Particularly, the method implements online processing, updating profiles and classification models on each incoming event. First, user profiles are built using creator-, content- and context-based features engineered through \textsc{nlp}. Then, unsupervised methods are exploited to create clusters of representative features. Finally, interpretable stream-based \textsc{ml} classifiers establish the trustworthiness of tweets in real-time. As a result, the proposed method provides the user with a dashboard, combining visual data and natural language knowledge, to make tweet classification transparent.

\section{Proposed method}
\label{sec:proposed_method}

\begin{figure}
 \centering
 \includegraphics[scale=0.12]{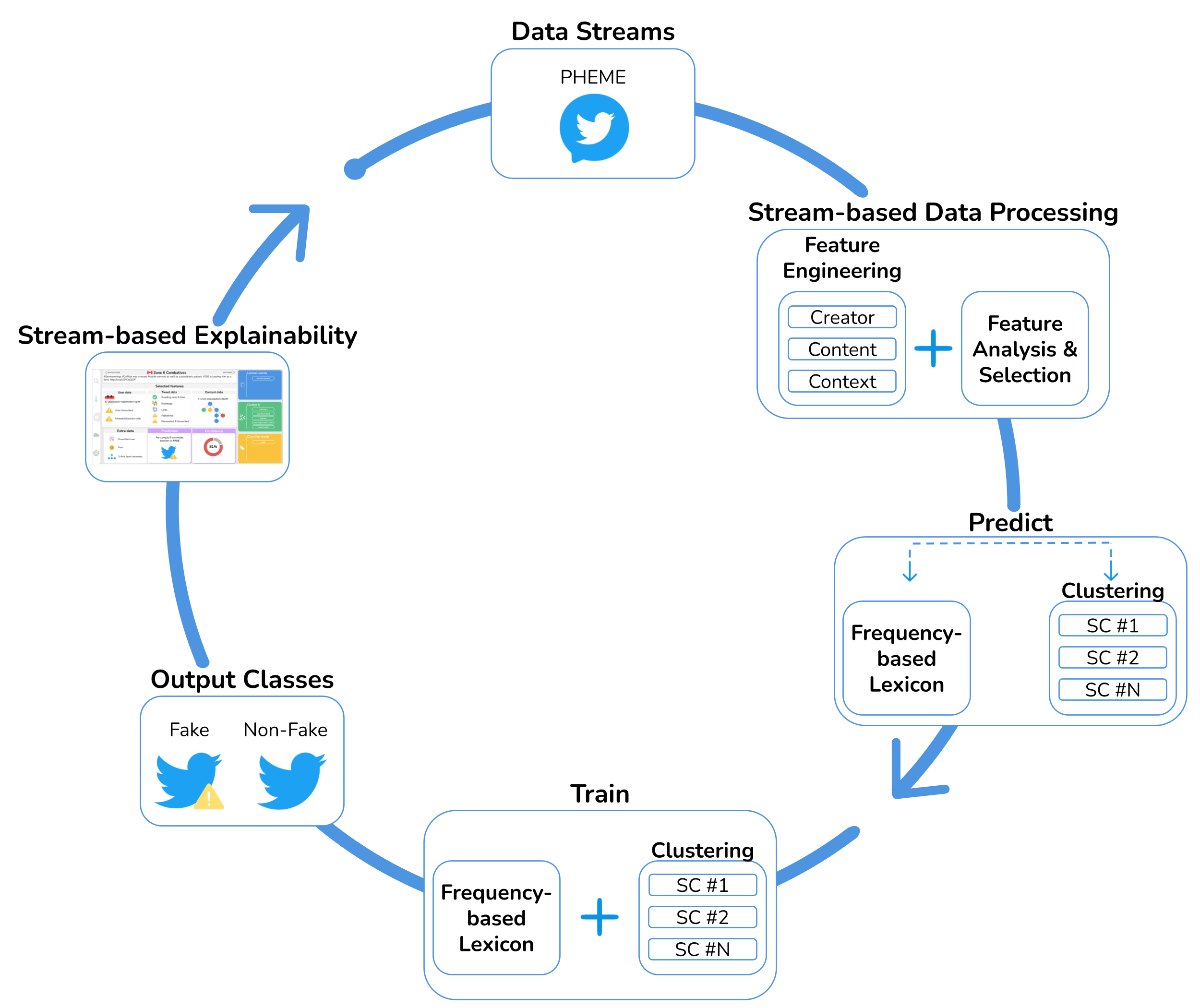}
 \caption{System diagram composed of: (\textit{i}) stream-based data processing, (\textit{ii}) online classification, and (\textit{iii}) stream-based explainability.}
 \label{fig:scheme}
\end{figure}

The proposed online and explainable fake news detection system is described in Figure \ref{fig:scheme}. It is composed of three main modules: (\textit{i}) the stream-based data processing module (Section \ref{sec:data_processing}) which comprises feature engineering (Section \ref{sec:feature_engineering}), and analysis and selection tasks (Section \ref{sec:feature_analysis_selection}); (\textit{ii}) the stream-based classification module (Section \ref{sec:classification}) composed of lexicon-based (Section \ref{sec:lexicon_classification}), unsupervised and supervised (Section \ref{sec:unsupervised_supervised_classification}) classifiers; and (\textit{iii}) the stream-based explainability module (Section \ref{sec:explainability}). The explored data comprises two collections of tweets related to breaking news events released in 2016 ({\sc pheme}) and augmented in 2018 ({\sc pheme-r}).

\subsection{Stream-based data processing}
\label{sec:data_processing}

This module exploits \textsc{nlp} techniques to take full advantage of the \textsc{ml} models. Firstly, the feature engineering process generates new knowledge from the experimental data. Then, it analyses the resulting feature set to finally select the most relevant features for the classification.

\subsubsection{Feature engineering}
\label{sec:feature_engineering}

The proposed system computes features from a wide spectrum: (\textit{i}) creator-, (\textit{ii}) content- (lexical and syntactical features, stylistic features, and visual features), and (\textit{iii}) context-based (network, distribution and temporal) features.

The creator-based features specify whether the user has an account description, a profile image and if the account has been protected and/or verified, the timezone, the number of followers and friends, the ratio between friends and followers, as well as the number of favourite tags received by the user. In the end, the time span in days between user registration and tweet post is calculated along with the weekly post frequency of the user\footnote{These last two features may be considered as context-based temporal.}.

The linguistic and syntactic content-based features include the word $n$-grams from the processed tweet and whether the content is duplicated in the experimental data set. The physical style features comprise the adjective, auxiliary, bad word, determiner, difficult word, hashtag, link (also repeated), noun, pronoun, punctuation, uppercase word and word counters. The sentiment-related features comprise emotion (anger, fear, happiness, sadness and surprise) and polarity (negative, neutral and positive). The non-physical style-based features are based on the Flesch reading ease metric (see Table \ref{tab:flesch_reading_ease}), the McAlpine \textsc{eflaw} readability score for English foreign speakers\footnote{It is recommended to be equal or lower than 25 points.} and the reading time in seconds. Concerning visual-based features, the system verifies if the tweet contains links to images and videos.

The generated context-based features consider whether the tweet has been retweeted and/or favourited, the depth of the retweet distribution network and the number of first-level retweets. Finally, the distribution pattern is analysed through the retweet and favourite counters. 

\begin{table}[!htbp]
\centering
\footnotesize
\caption{Flesch reading ease score and difficulty.}
\label{tab:flesch_reading_ease}
\begin{tabular}{ll}
\toprule
\bf Score & \bf Difficulty\\
\midrule
90-100& Very Easy\\
80-89 & Easy\\
70-79 & Fairly Easy\\
60-69 & Standard\\
50-59 & Fairly Difficult\\
30-49 & Difficult\\
0-29 & Very Confusing\\\bottomrule
\end{tabular}
\end{table}

The specific techniques applied to compute the aforementioned features will be described in Section \ref{sec:feature_engineering_results} along with the data processing details.

\subsubsection{Feature analysis and selection}
\label{sec:feature_analysis_selection}

Prior to feature selection, the system computes the variance of the features to establish their relative importance and, finally, discard those with low variance. Thus, the feature space dimension is reduced to minimize the computational load and time needed by \textsc{ml} models to classify tweets.

\subsection{Stream-based classification}
\label{sec:classification}

The proposed method involves lexicon-based (Section \ref{sec:lexicon_classification}), unsupervised and supervised classification (\ref{sec:unsupervised_supervised_classification}) in both the predict and train steps of each incoming tweet.

\subsubsection{Frequency-based lexicon}
\label{sec:lexicon_classification}

The adopted frequency-based lexicon is applied to the content of each incoming tweet. Algorithm \ref{alg:lexicon} provides the corresponding pseudo-code. The lexica allow swift prediction followed by updating (training) based on the tweet content. The training stage considers the target class the $n$-grams represent and their frequency. More in detail, it defines three thresholds: (\textit{i}) the $n$-gram range to extract the words; (\textit{ii}) the number of elements to be included in the resulting lexica; and (\textit{iii}) the frequency used as insert condition.

\begin{algorithm}
\caption{Frequency-based lexicon generation}\label{alg:lexicon}
\begin{algorithmic}[1]
\footnotesize
\Require Content-related data of the incoming tweet: \texttt{tweet\_processed, ngram\_range, num\_elements, threshold, frequency\_lexicon.}

\Ensure The algorithm returns the \texttt{fake\_lexicon} and \texttt{non-fake\_lexicon} disjoint sets.

\State tweet\_ngrams=$generate\_ngrams($tweet\_processed, ngram\_range); // \texttt{Holds the $n$-gram representation (within the given $n$-gram\_range of the tweet content) and frequency.}

\State frequency\_lexicon=$update\_frequency\_lexicon($frequency\_lexicon, tweet\_ngrams);

\State fake\_lexicon=$frequency\_lexicon$[$class=$ fake, $frequency$\textbf{ > }threshold][0:num\_elements];
 
\State non-fake\_lexicon=$frequency\_lexicon$[$class=$ non-fake, $frequency$\textbf{ > }threshold][0:num\_elements];

return $frequency\_lexicon$, $fake\_lexicon$, $non$-$fake\_lexicon$. // \texttt{Returns the updated frequency\_lexicon plus the fake and non-fake lexica.}

\end{algorithmic}
\end{algorithm}

\subsubsection{Unsupervised and supervised classification}
\label{sec:unsupervised_supervised_classification}

First, the unsupervised classification creates clusters of comparable spatial extent, by splitting the input data based on their proximity. It applies \textit{k}‐means clustering \citep{Sinaga2020,Vouros2021} to minimise within-cluster variances, also known as squared Euclidean distances. Then, for each discovered cluster, one supervised classifier is trained.

The method involves several well-known stream-based \textsc{ml} models, selected according to their good performance in similar classification problems \citep{Aphiwongsophon2018,Xiao2020,Caio2021}.

\begin{itemize}
 
 \item \textbf{Adaptive Random Forest Classifier} (\textsc{arfc}) \citep{Gomes2017}. It induces diversity using re-sampling, random feature subsets for node splits and drift detectors per base tree.
 
 \item \textbf{Hoeffding Adaptive Tree Classifier} (\textsc{hatc}) \citep{Bifet2009}. It uses a drift detector to monitor branch performance. Moreover, it presents a more efficient and effective bootstrap sampling strategy compared to the original Hoeffding Tree classifier.
 
 \item \textbf{Hoeffding Tree Classifier} (\textsc{htc}) \citep{Pham2017}. It is an incremental decision tree algorithm which quantifies the number of samples needed to estimate the statistics while guarantying the prescribed performance.
 
 \item \textbf{Gaussian Naive Bayes} (\textsc{gnb}) \citep{Xue2021}. It enhances the original Naive Bayes algorithm by exploiting a Gaussian distribution per feature and class. 
\end{itemize}

Algorithmic performance is determined with the help of classification accuracy, \textit{F}-measure (macro and micro-averaging) and run-time metrics, following the prequential evaluation protocol \citep{Gama2013}.

\subsection{Stream-based explainability module}
\label{sec:explainability}

Transparency is essential to make results both understandable and trustworthy for the end users. This means that outcomes need to be accompanied by explanatory descriptions. The designed fake news classification solution relies on interpretable models to obtain and present the relevant data in an explainability dashboard. The explanation of each prediction includes:
\begin{itemize}
 \item Relevant user, content and context features selected by the supervised \textsc{ml} models.
 \item Predicted class (fake and non-fake) together with confidence.
 \item $K$ disjoint elements ordered by their appearance frequency extracted from the fake and non-fake lexica.
 \item $K$ features that surround the centroid of the cluster to which the entry belongs.
\end{itemize}

The latter is completed with natural language descriptions of the corresponding tree decision path.

\section{Experimental results}
\label{sec:experimental_results}

All experiments were performed using a server with the following hardware specifications:
\begin{itemize}
 \item Operating System: Ubuntu 18.04.2 LTS 64 bits
 \item Processor: Intel\@Core i9-10900K \SI{2.80}{\giga\hertz}
 \item RAM: \SI{96}{\giga\byte} DDR4 
 \item Disk: \SI{480}{\giga\byte} NVME + \SI{500}{\giga\byte} SSD
\end{itemize}

\subsection{Experimental data sets}
\label{sec:experimental_dataset}

The experiments were performed with temporally ordered data streams created from the \textsc{pheme} and \textsc{pheme-r} data sets\footnote{Available at \url{https://figshare.com/articles/dataset/PHEME_dataset_for_Rumour_Detection_and_Veracity_Classification/6392078} and \url{https://figshare.com/articles/dataset/PHEME_dataset_of_rumours_and_non-rumours/4010619}, June 2023.} and, for additional testing, from the \cite{Nikiforos2020} data set\footnote{Available at \url{https://hilab.di.ionio.gr/wp-content/uploads/2020/02/HILab-Fake_News_Detection_For_Hong_Kong_Tweets.xlsx}, June 2023.}. The \textsc{pheme} collections comprise \num{6424} tweets created by \num{2893} users between August 2014 and March 2015. All tweets were manually labeled as fake and non-fake. The data set from \cite{Nikiforos2020} contains \num{2366} tweets posted by \num{51} users between April 2013 and December 2019. This data set was exclusively used to confirm the performance of the proposed method (see Section \ref{sec:discussion}). Table \ref{tab:dataset} details the number of users and tweets per class in each experimental data set.

\begin{table}[!htbp]
\centering
\footnotesize
\caption{\label{tab:dataset}Classes, number of users and tweets of the experimental data sets.}
\begin{tabular}{llS[table-format=4]S[table-format=4]}
\toprule
\bf Data set & \bf Class & {\bf Users} & {\bf Tweets}\\
\midrule
\multirow{3}{*}{\textsc{pheme}} & Fake & 1023 & 2402\\
& Non-fake & 2204 & 4022\\\cmidrule{2-4}
& Total & 2893 & 6424\\\midrule

\multirow{3}{*}{\cite{Nikiforos2020}} & Fake & 42 & 272 \\
& Non-fake & 9 & 2094\\\cmidrule{2-4}
& Total & 51 & 2366\\

\bottomrule
\end{tabular}
\end{table}

\subsection{Stream-based data processing}
\label{sec:data_processing_results}

As previously mentioned, data processing applies \textsc{nlp} techniques to ensure the competing performance of the \textsc{ml} models. The procedures used for online feature engineering, analysis and selection are presented below.

\subsubsection{Feature engineering}
\label{sec:feature_engineering_results}

Firstly, tweet content is purged from \textsc{url}, redundant blank spaces, special characters (non-alphanumerical items, like accents and punctuation marks) and stop-words from the list provided by the Natural Language Toolkit (\textsc{nltk})\footnote{Available at \url{https://gist.github.com/sebleier/554280}, June 2023.}. The remaining content is lemmatised with the English \texttt{en\_core\_web\_md} model\footnote{Available at \url{https://spacy.io/models/en}, June 2023.} of the spaCy library\footnote{Available at \url{https://spacy.io}, June 2023.} and content polarity is established with \texttt{TextBlob}\footnote{Available at \url{https://pypi.org/project/spacytextblob}, June 2023.}, a sentiment analysis component for spaCy. The tweet emotion is calculated using \texttt{Text2emotion} Python library\footnote{Available at \url{https://pypi.org/project/text2emotion}, June 2023.}.

The creation of non-physical style features relies on the \texttt{TextDescriptives}\footnote{Available at \url{https://spacy.io/universe/project/textdescriptives}, June 2023.} spaCy module (features 13, 14, 17, 26, 28 and 29 in Table \ref{tab:features}) and on the \texttt{Textstat}\footnote{Available at \url{https://pypi.org/project/textstat}, June 2023.} Python library (features 18, 20, 25 and 30 in Table \ref{tab:features}). The bad word count (feature 15 in Table \ref{tab:features}) depends on the list provided by Wikimedia Meta-wiki\footnote{Available at \url{https://meta.wikimedia.org/wiki/Research:Revision_scoring_as_a_service/Word_lists/en}, June 2023.}.

Given the importance of hashtags within tweets, hashtags are decomposed into their elementary constituents, \textit{i.e.}, words. This is applied to the cases where the hashtag is not represented in title format\footnote{The first letter of each of the words which compose the hashtag capitalised}. This splitter uses a freely available English corpus, the \texttt{Alpha} lexicon\footnote{Available at \url{https://github.com/dwyl/english-words}, June 2023.}, along with the English corpus by \cite{Garcia-Mendez2019}. It employs a recursive and reentrant algorithm to minimise the number of splits needed to decompose the hashtag into correct English words. As an example, the proposed text decomposition solution splits \textit{hatecannotdriveouthate} as \textit{hate cannot drive out hate}.

The word $n$-grams are extracted from the accumulated tweet textual data using \texttt{CountVectorizer}\footnote{Available at \url{https://scikit-learn.org/stable/modules/generated/sklearn.feature_extraction.text.CountVectorizer.html}, June 2023.} Python library. Listing \ref{ngrams_conf} shows the ranges and best values for the \texttt{CountVectorizer} configuration parameters based on iterative experimental tests with \texttt{GridSearch}\footnote{Available at \url{https://scikit-learn.org/stable/modules/generated/sklearn.model_selection.GridSearchCV.html}, June 2023.} meta transformer wrapper for the \textsc{hatc} classifier.

\begin{lstlisting}[frame=single,caption={Parameter ranges for the generation of $n$-grams (best values in bold).}, label={ngrams_conf},escapechar=ä]
maxdf = [ä\bf 0.7ä, 0.5, 0.3]
mindf = [0.1, ä\bf 0.01ä, 0.001]
ngramrange = [(1, 2), ä\bf (1, 3)ä, (1,4)]
\end{lstlisting}

Table \ref{tab:features} shows the creator-, content- and context-based features selected for the detection of fake news. An additional pair of features is created for each user and numerical feature in Table \ref{tab:features} (features 6-9, 13-18, 21-24, 26, 28, 29, 31-33, 39 and 40): the user incremental feature average and latest feature trend, a Boolean feature that compares the last user feature value with the current user feature average\footnote{True if the feature value is equal or higher than the user feature average; otherwise is false.}.

\begin{table}[!htbp]
\centering
\footnotesize
\caption{\label{tab:features}Features considered for the classification by profile (creator, content, context) and data type (Boolean, categorical, numerical, textual).}
\begin{tabular}{llcl} 
\toprule
\bf Profiling & \bf Data type & \bf Number & \bf Name \\\midrule
\multirow{14}{*}{Creator-based}& \multirow{4}{*}{Boolean} & 1 & Has profile description\\
& & 2 & Has profile image\\
& & 3 & Protected\\
& & 4 & Verified\\\cmidrule{2-4}
& Categorical & 5 & Timezone\\\cmidrule{2-4}
& \multirow{5}{*}{Numerical} & 6 & Follower count\\
& & 7 & Friend count\\
& & 8 & Friends-followers ratio\\
& & 9 & User favourite count\\
& & 10 & \makecell[l]{Tweet-registration \\time spam (in days)}\\
& & 11 & Weekly tweet frequency\\
\midrule

& Boolean & 12 & Text duplicated\\
\cmidrule{2-4}
\multirow{24}{*}{Content-based}& \multirow{20}{*}{Numerical} & 13 & Adjective count\\
& & 14 & Auxiliary count\\
& & 15 & Bad word count\\
& & 16 & Char count\\
& & 17 & Determiner count\\
& & 18 & Difficult word count\\
& & 19 & \makecell[l]{Emotion (anger, fear, \\ happiness, sadness, surprise)}\\
& & 20 & Flesch reading ease\\
& & 21 & Hashtag count\\
& & 22 & Image count\\
& & 23 & Link count\\
& & 24 & Link repeated count\\
& & 25 & McAlpine \textsc{eflaw} readability\\
& & 26 & Noun count\\
& & 27 & Polarity\\
& & 28 & Pronoun count\\
& & 29 & Punctuation count\\
& & 30 & Reading time\\
& & 31 & Uppercase word count\\
& & 32 & Video count\\
& & 33 & Word count\\\cmidrule{2-4}

& Textual & 34 & Word $n$-grams\\\midrule

\multirow{7}{*}{Context-based} & \multirow{2}{*}{Boolean} & 35 & Retweeted\\
& & 36 & Tweet favourited\\\cmidrule{2-4}

& \multirow{4}{*}{Numerical} & 37 & Distribution depth\\ 
& & 38 & First level retweet \\
& & 39 & Retweet count\\
& & 40 & Tweet favourite count\\
\bottomrule
\end{tabular}
\end{table}

\subsubsection{Feature analysis and selection}
\label{sec:feature_analysis_selection_results}

The method analyses the variance of features in Table \ref{tab:features} to compute their relative importance. Those features with low variance are discarded. Particularly, feature selection is performed at each incoming event using the \texttt{VarianceThreshold}\footnote{Available at \url{https://riverml.xyz/0.11.1/api/feature-selection/VarianceThreshold}, June 2023.} algorithm from River\footnote{Available at \url{https://riverml.xyz/0.11.1}, June 2023.} library to improve the fake class recall metric.

\subsection{Stream-based classification}
\label{sec:classification_results}

Online classification involves prediction and training for each incoming sample. This section presents the results obtained by the lexicon-based, unsupervised and supervised classification procedures.

\subsubsection{Frequency-based lexicon}
\label{sec:lexicon_classification_results}

The building of dynamic frequency-based lexicon starts after accumulating \SI{5}{\percent} of the samples. More in detail, the system extracts \num{700} from 2- to 4-word-length unique elements for each target class (fake and non-fake). Listing \ref{lexicon_conf} provides the configuration parameter ranges. Best values were obtained once again from iterative experimental tests and using the \textsc{hatc} classifier.

\begin{lstlisting}[frame=single,caption={Parameter ranges for the generation of the frequency-based lexicon (best values in bold).}, label={lexicon_conf},emphstyle=\textbf,escapechar=ä]
ngrams = [(1,4),ä\bf (2,4)ä,(3,4)]
numberwords = [800,ä\bf 700ä,600,500, 400, 300, 200]
minfreqvalue = [ä\bf 1ä, 2, 5, 8, 10, 15, 20, 30]
\end{lstlisting}

\subsubsection{Unsupervised and supervised classification results}
\label{sec:unsupervised_supervised_classification_results}

As described in Section \ref{sec:classification}, the first step applies unsupervised clustering. The latter uses the widely known \textit{k}-means model\footnote{Available at \url{https://riverml.xyz/dev/api/cluster/KMeans}, June 2023.}. Then, for each of the discovered clusters, one supervised classifier is trained using the following implementations:

\begin{itemize}

 \item \textsc{arfc}\footnote{Available at \url{https://riverml.xyz/0.11.1/api/ensemble/AdaptiveRandomForestClassifier}, June 2023.}

 \item \textsc{hatc}\footnote{Available at \url{https://riverml.xyz/0.11.1/api/tree/HoeffdingAdaptiveTreeClassifier}, June 2023.}

 \item \textsc{htc}\footnote{Available at \url{https://riverml.xyz/0.11.1/api/tree/HoeffdingTreeClassifier}, June 2023.}

 \item \textsc{gnb}\footnote{Available at \url{https://riverml.xyz/0.11.1/api/naive-bayes/GaussianNB}, June 2023.}

\end{itemize}

Hyperparameter optimisation is performed for the aforementioned \textsc{ml} algorithms. Listings \ref{arfc_conf}, \ref{hatc_conf}, \ref{htc_conf} and \ref{gnb_conf} show the configuration ranges and best values (in bold) for each algorithm.

\begin{lstlisting}[frame=single,caption={Hyperparameter ranges for the \textsc{arfc} model (best values in bold).},label={arfc_conf},emphstyle=\textbf,escapechar=ä]
clusters = [ä\bf 10ä, 20, 30]
models = [50, 100, ä\bf 200ä]
features = [ä\bf 50ä, 100, 200]
lambda = [ä\bf 50ä, 100, 200]
\end{lstlisting}

\begin{lstlisting}[frame=single,caption={Hyperparameter ranges for the \textsc{hatc} model (best values in bold).},label={hatc_conf},emphstyle=\textbf,escapechar=ä]
clusters = [ä\bf 10ä, 20, 30]
depth = [ä\bf 50ä, 100, 200]
tiethreshold = [ä\bf 0.5ä, 0.05, 0.005]
maxsize = [50, 100, ä\bf 200ä]
\end{lstlisting}

\begin{lstlisting}[frame=single,caption={Hyperparameter ranges for the \textsc{htc} model (best values in bold).},label={htc_conf},emphstyle=\textbf,escapechar=ä] 
clusters = [ä\bf 10ä, 20, 30]
depth = [ä\bf 50ä, 100, 200]
tiethreshold = [ä\bf 0.5ä, 0.05, 0.005]
maxsize = [ä\bf 50ä, 100, 200]
\end{lstlisting}

\begin{lstlisting}[frame=single,caption={Hyperparameter ranges for the \textsc{gnb} model (best value in bold).},label={gnb_conf},emphstyle=\textbf,escapechar=ä]
clusters = [ä\bf 10ä, 20, 30]
\end{lstlisting}

Table \ref{tab:classification_results} shows the performance of the \textsc{ml} models. Set \textsc{a} of features includes those in Table \ref{tab:features} except for word $n$-grams, whereas, set \textsc{b} includes set \textsc{a} plus the latter textual features. Finally, set \textsc{c} is composed of set \textsc{b} plus the frequency-based lexicon. The proposed solution exhibits a processing time of \SI{0.42}{\second\per sample} in the worst scenario (\textsc{arfc} model and the set of features \textsc{a}), which can be considered real time.

In light of the results, \textsc{arfc} exhibits the best performance with all feature sets and for all evaluation metrics. The use of word $n$-grams results in significant improvement across all algorithms. The highest boost occurs for the \textsc{gnb} model (+\SI{12}{\percent} percent points in accuracy and micro \textit{F}-measure for the fake class). Despite the promising results, micro \textit{F}-measure values for the target fake class remain under the \SI{70}{\percent} threshold with feature sets \textsc{a} and \textsc{b}. Finally, the solution reaches accuracy and macro \textit{F}-measure about \SI{80}{\percent} with all engineered features (set \textsc{c}).

\begin{table}[!htbp]
\centering
\footnotesize
\caption{\label{tab:classification_results}Online fake detection results in terms of accuracy, macro and micro \textit{F}-measure (best values in bold) and run-time for the \textsc{arfc}, \textsc{hatc}, \textsc{htc} and \textsc{gnb} models by feature set.}
\begin{tabular}{clccccS[table-format=6.2]}
\toprule
\bf {Set} & \bf {Classifier} & \bf {Accuracy} & \multicolumn{3}{c}{\bf \textit{F}-measure} & {\bf Time}\\
\cmidrule(lr){4-6}
 & & & Macro & \#non-fake & \#fake & {(s)}\\
\midrule
\multirow{4}{*}{A} & 
\textsc{~~~arfc} & \bf 73.09 & \bf 70.62 & \bf 79.14 & \bf 62.10 & 2677.82\\&
\textsc{~~~hatc} & 64.95 & 63.29 & 71.10 & 55.49 & ~~~~8.91\\&
\textsc{~~~htc} & 64.76 & 62.79 & 71.36 & 54.21 & ~~~~7.45\\&
\textsc{~~~gnb} & 52.95 & 49.29 & 62.91 & 35.68 & ~~~~6.36\\
\midrule

\multirow{4}{*}{B} & 
\textsc{~~~arfc} & \bf 75.43 & \bf 73.17 & \bf 80.96 & \bf 65.38 & 1644.25\\&
\textsc{~~~hatc} & 70.11 & 66.28 & 77.65 & 54.90 & ~~~29.88\\&
\textsc{~~~htc} & 69.46 & 64.59 & 77.72 & 51.47 & ~~~23.25\\&
\textsc{~~~gnb} & 64.09 & 60.21 & 72.64 & 47.79 & ~~~20.44\\
\midrule

\multirow{4}{*}{C} & 
\textsc{~~~arfc} & \bf 80.26 & \bf 78.97 & \bf 84.18 & \bf 73.77 & 1910.07\\& 
\textsc{~~~hatc} & 78.20 & 76.42 & 82.91 & 69.92 & ~299.35\\& 
\textsc{~~~htc} & 77.94 & 76.11 & 82.72 & 69.51 & ~293.74\\& 
\textsc{~~~gnb} & 74.66 & 73.45 & 79.13 & 67.76 & ~286.24\\
\bottomrule
\end{tabular}
\end{table}

\subsubsection{Discussion}
\label{sec:discussion}

Since the majority of the competing works implement batch rather than stream processing and use different data sets, result comparison may not be straightforward. Batch and stream results are only directly comparable if obtained with the same data samples. This means that, ideally, the comparison should be made with a chronologically ordered data set, and the evaluation should consider only the test partition samples. In the case of stream processing, this is achieved by setting the dimension of the sliding window to the number of samples of the test partition and then processing the data set as a stream. 

The batch classification works by \cite{Zubiaga2017}, \cite{Akinyemi2020} and \cite{Ying2021} explore the same \textsc{pheme} data set with cross-folded validation, using \SI{80}{\percent} of the samples for training and \SI{20}{\percent} for testing. The related online fake news classification system of \cite{Ksieniewicz2020} employs another data set, preventing direct comparison.

Table \ref{tab:comparison_results} provides the theoretical comparison results of the most related works together with those of the proposed solution with a sliding window holding \SI{20}{\percent} of the data (for offline comparison) and a sliding window comprising all data (for online comparison)\footnote{\textsc{na} is used to indicate when the competing works did not provide results for specific metrics.}. The proposed solution with a sliding window of \SI{20}{\percent} of the data achieves an improvement in macro \textit{F}-measure of \num{20.12} and \num{17.42} percent points with respect to the work of \cite{Zubiaga2017} and \cite{Jain2022}, respectively. Moreover, it attains \num{+4.62} percent points in fake \textit{F}-measure compared to \cite{Akinyemi2020}. When compared with the batch and online deep learning approaches of \cite{Ying2021} and \cite{Ksieniewicz2020}, the proposed solution exhibits slightly lower performance but grants algorithmic transparency with lesser memory and computation time. Finally, for a fair comparison with the most related work by \cite{Ksieniewicz2020}, due to the fact the authors provided the implementation of the solution, we were able to run the experiments with the \textsc{pheme} data set and the accuracy obtained in this regard is \SI{74.10}{\percent} (\num{-6.16} percent points than our proposal).

\begin{table}[!htbp]
\centering
\footnotesize
\caption{\label{tab:comparison_results}Fake detection theoretical comparison in terms of accuracy, macro and micro \textit{F}-measure between related works and the proposed solution.}
\begin{threeparttable}
 \begin{tabular}{l@{}cccccS[table-format=6.2]}
 \toprule
 \bf Authorship & \bf Processing & \bf {Accuracy} & \multicolumn{3}{c}{\bf \textit{F}-measure}\\
 \cmidrule(lr){4-6}
 & & & Macro & \#non-fake & \#fake\\
 \midrule
 
 \cite{Zubiaga2017} & Offline & \textsc{na}\tnote{1} & 60.70 & \textsc{na} & \textsc{na}\\
 
 \cite{Akinyemi2020} & Offline & 81.90 & 78.00 & 87.00 & 70.00\\
 
 \cite{Ying2021} & Offline & 87.20 & \textsc{na} & 90.40 & 80.70\\

 \cite{Jain2022} & Offline & \textsc{na} & 63.40&\textsc{na} &\textsc{na}\\
 \midrule
 
 \cite{Ksieniewicz2020} & Online & 81.90 & \textsc{na} & \textsc{na} & \textsc{na}\\
 \midrule
 \multirow{2}{*}{\bf Proposed solution} & Online\tnote{2} & 82.82 & 80.82 & 87.02 & 74.62\\
 & Online\tnote{3} & 80.26 & 78.97 & 84.18 & 73.77\\
 & Offline\tnote{4} & 99.14 & 97.54 & 99.52 & 95.56\\
 \bottomrule
 \end{tabular}
 \begin{tablenotes}
 \item[1] Not available. 
 \item[2] Sliding window holds \SI{20}{\percent} of data.
 \item[3] Sliding window holds the full data.
 \item [4] Sliding window holds \SI{20}{\percent} of data for the data set provided by \cite{Nikiforos2020}.
 \end{tablenotes}
\end{threeparttable}
\end{table}

Originally, \cite{Nikiforos2020} achieved an accuracy of \SI{99.79}{\percent} and \SI{99.37}{\percent} with Naive Bayes and \textsc{rf} offline classifiers, respectively. Both models were trained with a synthetic minority over-sampled set generated from \SI{80}{\percent} of the original data (to overcome the class imbalance of the original data) and tested with the \SI{20}{\percent} of the original data. To compare with these results, the experiment was repeated with a sliding window comprising \SI{20}{\percent} of the total number of samples and the best \textsc{arfc} model. In this case, the current solution attained \SI{99.14}{\percent} accuracy, macro \textit{F}-measure of \SI{97.54}{\percent}, and micro \textit{F}-measure of \SI{99.52}{\percent} and \SI{95.56}{\percent} for non-fake and fake classes, respectively. This means that the proposed online method achieves, without oversampling and in real time, a comparable accuracy.

\subsection{Stream-based explainability module}
\label{sec:explainability_results}

\begin{figure}
\centering
\begin{subfigure}{0.7\textwidth}
 \includegraphics[width=\textwidth]{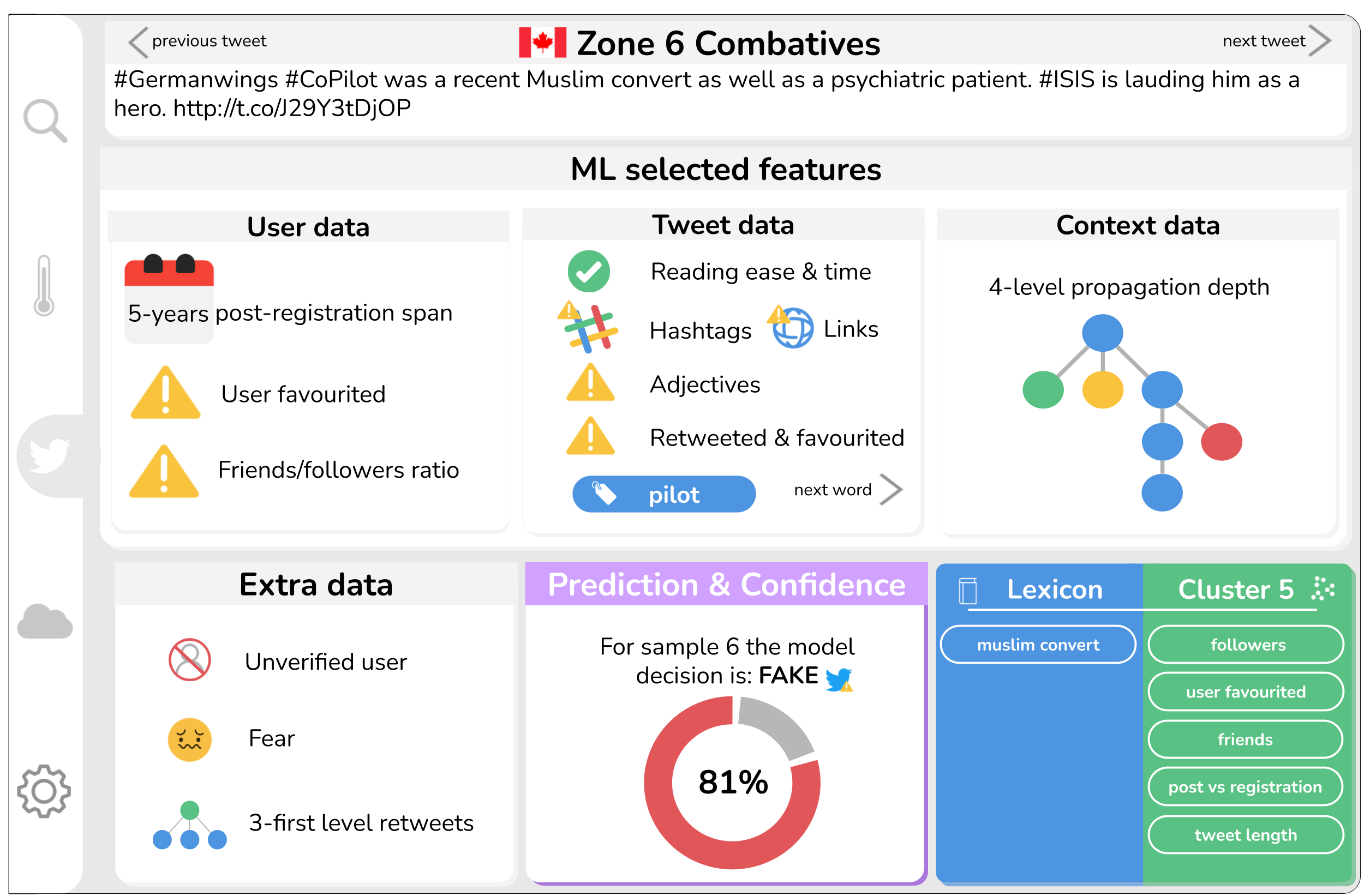}
\end{subfigure}
\hfill
\begin{subfigure}{0.7\textwidth}
 \includegraphics[width=\textwidth]{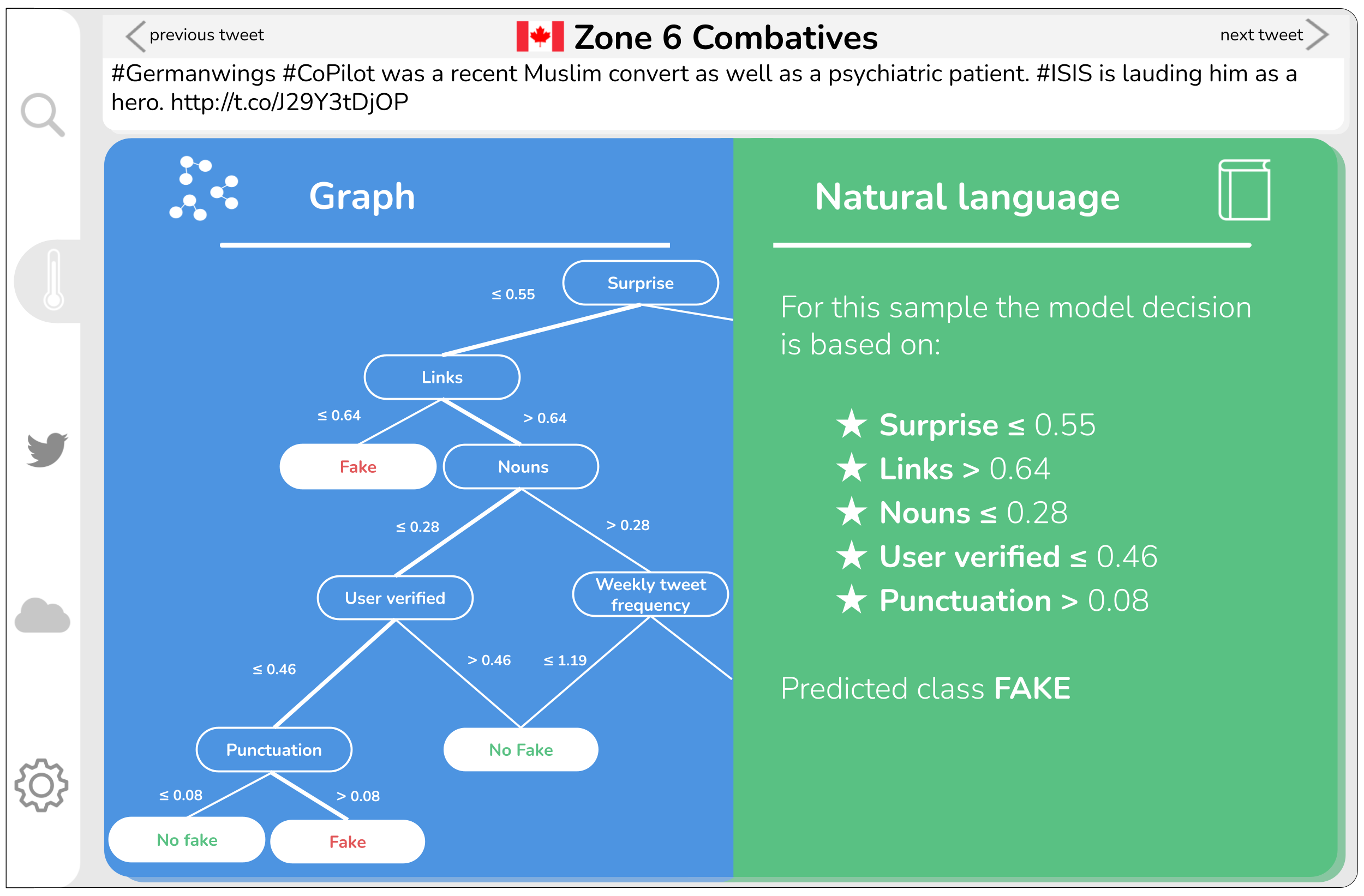}
\end{subfigure}
\caption{Explainability dashboard comprising: (\textit{i}) selected features from the content, context, and creator, (\textit{ii}) the prediction, (\textit{iii}) representative entries of the frequency-based lexicon and the clustering procedure, and (\textit{iv}) the decision path and its natural language transcription.}
\label{fig:dashboard}
\end{figure}

Figure \ref{fig:dashboard} shows the user explainability dashboard, which aims to make the model outcome comprehensible. In the upper part, it displays the classification of the tweet sample. The user name is \textit{Zone 6 Combatives} and the timezone Canadian. The top center displays the tweet content and the center presents the creator-, content- and context-related features selected by the \textsc{ml} classifier. Feature warnings are shown when a feature deviates from the user average as is the case of \textit{reading ease \& time} feature. Otherwise, the features include an \textsc{ok} symbol as in the case of the \textit{5-years post-registration span} feature. The classifier singled out the word \textit{pilot} as relevant. The tweet was classified as fake with an \SI{81}{\percent} of confidence, according to the \texttt{Predict\_Proba\_One}\footnote{Available at \url{https://riverml.xyz/0.11.1/api/base/Classifier}, June 2023.} from River \textsc{ml} library. In the end, the most representative features for both the frequency-based lexicon and the clustering procedure are provided\footnote{The sample belongs to cluster 5.}. 

The bottom part of the dashboard displays the decision tree path (obtained using \texttt{debug one} and \texttt{draw}\footnote{Available at \url{https://riverml.xyz/0.11.1/api/tree/HoeffdingAdaptiveTreeRegressor}, June 2023.} libraries) and the corresponding natural language description. Particularly, the first decision is based on the \texttt{surprise} feature (see feature 19 in Table \ref{tab:features}). If its value is lower or equal to 0.55, the reasoning continues through the left branch. Otherwise it goes to the right branch.

\section{Conclusion}
\label{sec:conclusion}

Social media is becoming an increasing source of breaking news. In these platforms, information is shared regardless of the context and reliability of the content and creator of the posted information. This instant news dissemination and consumption model easily propagates fake news, constituting a challenge in terms of transparency, reliability, and real-time processing. Accordingly, the proposed solution addresses transparency through explanations, reliability through fake news detection, and real-time processing through incremental profiling and learning. The motivation for the current work relies on the early detection, isolation and explanation of misinformation, all of them crucial procedures to increase the quality and trust in digital media social platforms.

More in detail, this work contributes with an explainable classification method to recognise fake news in real-time. The proposed method combines both unsupervised and supervised approaches with online created lexica. Specifically, it comprises (\textit{i}) stream-based data processing (through feature engineering, analysis and selection), (\textit{ii}) stream-based classification (lexicon-based, unsupervised and supervised classification), and (\textit{iii}) stream-based explainability (prediction confidence and interpretable classification). Furthermore, the profiles are built using creator-, content- and context-based features with the help of \textsc{nlp} techniques. The experimental classification results of \SI{80}{\percent} accuracy and macro \textit{F}-measure, obtained with a real data set manually annotated, endorse the promising performance of the designed explainable real-time fake news detection method.

Analyzing the related work, this proposal is the first to jointly provide stream-based data processing, profiling, classification and explainability. Future work will attempt to mitigate further the impact of fake news within social media by automatically identifying and isolating potential malicious accounts as well as extend the research to related tasks like stance detection, by exploiting new creator-, content- and context-based features.

\section*{Declarations}

\subsection*{Funding}

This work was partially supported by: (\textit{i}) Xunta de Galicia grants ED481B-2021-118 and ED481B-2022-093, Spain; (\textit{ii}) Portuguese national funds through FCT –- Fundação para a Ciência e a Tecnologia (Portuguese Foundation for Science and Technology) -- as part of project UIDP/50014/2020 (DOI: 10.54499/UIDP/50014/2020  $\mid$ \url{https://doi.org/10.54499/UIDP/50014/2020}); and (\textit{iii}) University of Vigo/CISUG for open access charge.

\subsection*{Competing interests}
The authors have no competing interests to declare that are relevant to the content of this article.

\subsection*{Ethics approval}

Not applicable

\subsection*{Consent to participate}

Not Applicable

\subsection*{Consent for publication}

Not Applicable

\subsection*{Availability of data and material}

The used data is openly available.

\subsection*{Code availability}

The code will become available at Github upon acceptance of the manuscript.

\subsection*{Authors' contributions}

\textbf{Francisco de Arriba-Pérez}: Conceptualization, Methodology, Software, Resources, Data Curation, Writing - Original Draft, Writing - Review \& Editing. \textbf{Silvia García-Méndez}: Conceptualization, Methodology, Resources, Data Curation, Writing - Original Draft, Writing - Review \& Editing.
\textbf{Fátima Leal}: Methodology, Data Curation, Writing - Original Draft, Writing - Review \& Editing. \textbf{Benedita Malheiro}: Conceptualization, Methodology, Validation, Writing - Review \& Editing, Supervision. \textbf{Juan Carlos Burguillo-Rial}: Conceptualization, Methodology, Validation, Writing - Review \& Editing, Supervision.

\bibliography{mybibfile}

\begin{thebibliography}{51}
\providecommand{\natexlab}[1]{#1}
\providecommand{\url}[1]{{#1}}
\providecommand{\urlprefix}{URL }
\providecommand{\doi}[1]{\url{https://doi.org/#1}}
\providecommand{\eprint}[2][]{\url{#2}}
 \bibcommenthead

\bibitem[{Akinyemi et~al(2020)Akinyemi, Adewusi, and Oyebade}]{Akinyemi2020}
Akinyemi B, Adewusi O, Oyebade A (2020) {An Improved Classification Model for
  Fake News Detection in Social Media}. International Journal of Information
  Technology and Computer Science 12(1):34--43. \doi{10.5815/ijitcs.2020.01.05}

\bibitem[{Aphiwongsophon and Chongstitvatana(2018)}]{Aphiwongsophon2018}
Aphiwongsophon S, Chongstitvatana P (2018) {Detecting Fake News with Machine
  Learning Method}. In: Proceedings of the International Conference on
  Electrical Engineering/Electronics, Computer, Telecommunications and
  Information Technology. IEEE, pp 528--531, \doi{10.1109/ECTICon.2018.8620051}

\bibitem[{Bifet and Gavald{\`{a}}(2009)}]{Bifet2009}
Bifet A, Gavald{\`{a}} R (2009) {Adaptive Learning from Evolving Data Streams}.
  In: Lecture Notes in Computer Science (including subseries Lecture Notes in
  Artificial Intelligence and Lecture Notes in Bioinformatics), vol 5772 LCNS.
  Springer, p 249--260, \doi{10.1007/978-3-642-03915-7_22}

\bibitem[{Bondielli and Marcelloni(2019)}]{Bondielli2019}
Bondielli A, Marcelloni F (2019) {A survey on fake news and rumour detection
  techniques}. Information Sciences 497:38--55. \doi{10.1016/j.ins.2019.05.035}

\bibitem[{Castillo et~al(2011)Castillo, Mendoza, and Poblete}]{Castillo:2011}
Castillo C, Mendoza M, Poblete B (2011) {Information credibility on twitter}.
  In: Proceedings of the International Conference on World Wide Web.
  Association for Computing Machinery, pp 675--684,
  \doi{10.1145/1963405.1963500}

\bibitem[{Chora{\'{s}} et~al(2021)Chora{\'{s}}, Demestichas, Gielczyk, Herrero,
  Ksieniewicz, Remoundou, Urda, and Wo'zniak}]{Choras:2021}
Chora{\'{s}} M, Demestichas K, Gielczyk A, et~al (2021) {Advanced Machine
  Learning techniques for fake news (online disinformation) detection A
  systematic mapping study}. Applied Soft Computing 101:107,050--107,064.
  \doi{10.1016j.asoc.2020.107050}

\bibitem[{Dong et~al(2020)Dong, Victor, and Qian}]{Dong2020}
Dong X, Victor U, Qian L (2020) {Two-Path Deep Semisupervised Learning for
  Timely Fake News Detection}. IEEE Transactions on Computational Social
  Systems 7(6):1386--1398. \doi{10.1109TCSS.2020.3027639}

\bibitem[{Dulac-Arnold et~al(2021)Dulac-Arnold, Levine, Mankowitz, Li,
  Paduraru, Gowal, and Hester}]{dulac2021challenges}
Dulac-Arnold G, Levine N, Mankowitz DJ, et~al (2021) {Challenges of real-world
  reinforcement learning: definitions, benchmarks and analysis}. Machine
  Learning 110(9):2419--2468. \doi{10.1007/s10994-021-05961-4}

\bibitem[{Galli et~al(2022)Galli, Masciari, Moscato, and Sperlí}]{Galli2022}
Galli A, Masciari E, Moscato V, et~al (2022) {A comprehensive Benchmark for
  fake news detection}. Journal of Intelligent Information Systems
  59(1):237--261. \doi{10.1007/s10844-021-00646-9}

\bibitem[{Gama et~al(2013)Gama, Sebastião, and Rodrigues}]{Gama2013}
Gama J, Sebastião R, Rodrigues PP (2013) {On Evaluating Stream Learning
  Algorithms}. Machine Learning 90(3):317--346. \doi{10.1007/s10994-012-5320-9}

\bibitem[{García-Méndez et~al(2019)García-Méndez, Fernández-Gavilanes,
  Costa-Montenegro, Juncal-Martínez, González-Castaño, and
  Reiter}]{Garcia-Mendez2019}
García-Méndez S, Fernández-Gavilanes M, Costa-Montenegro E, et~al (2019) {A
  System for Automatic English Text Expansion}. IEEE Access 7:123,320--123,333.
  \doi{10.1109/ACCESS.2019.2937505}

\bibitem[{Goindani and Neville(2019)}]{Goindani2019}
Goindani M, Neville J (2019) {Social reinforcement learning to combat fake news
  spread}. In: Proceedings of the Conference on Uncertainty in Artificial
  Intelligence. Association for Uncertainty in Artificial Intelligence, pp
  1006--1016

\bibitem[{Gomes et~al(2017)Gomes, Bifet, Read, Barddal, Enembreck, Pfharinger,
  Holmes, and Abdessalem}]{Gomes2017}
Gomes HM, Bifet A, Read J, et~al (2017) {Adaptive random forests for evolving
  data stream classification}. Machine Learning 106(9-10):1469--1495.
  \doi{10.1007/s10994-017-5642-8}

\bibitem[{Hu et~al(2014)Hu, Xu, Liu, Mei, Chen, and Luo}]{Hu2014IEEE}
Hu C, Xu Z, Liu Y, et~al (2014) {Semantic Link Network-Based Model for
  Organizing Multimedia Big Data}. IEEE Transactions on Emerging Topics in
  Computing 2(3):376--387. \doi{10.1109/TETC.2014.2316525}

\bibitem[{Jain et~al(2022)Jain, Kumar, and Shrivastava}]{Jain2022}
Jain DK, Kumar A, Shrivastava A (2022) {CanarDeep: a hybrid deep neural model
  with mixed fusion for rumour detection in social data streams}. Neural
  Computing and Applications 34:15,129--15,140.
  \doi{10.1007/s00521-021-06743-8}

\bibitem[{Jang et~al(2021)Jang, Park, Lee, and Seo}]{Jang2021}
Jang Y, Park CH, Lee DG, et~al (2021) {Fake News Detection on Social Media A
  Temporal-Based Approach}. Computers, Materials \& Continua 69(3):3563--3579.
  \doi{10.32604cmc.2021.018901}

\bibitem[{Kozik et~al(2022)Kozik, Kula, Chora{\'{s}}, and
  Wo{\'{z}}niak}]{Kozik:2022}
Kozik R, Kula S, Chora{\'{s}} M, et~al (2022) {Technical solution to counter
  potential crime: Text analysis to detect fake news and disinformation}.
  Journal of Computational Science 60:101,576--101,582.
  \doi{10.1016/j.jocs.2022.101576}

\bibitem[{Ksieniewicz et~al(2020)Ksieniewicz, Zyblewski, Chora{\'{s}}, Kozik,
  Gielczyk, and Wozniak}]{Ksieniewicz2020}
Ksieniewicz P, Zyblewski P, Chora{\'{s}} M, et~al (2020) {Fake News Detection
  from Data Streams}. In: Proceedings of the International Joint Conference on
  Neural Networks. IEEE, pp 1--8, \doi{10.1109IJCNN48605.2020.9207498}

\bibitem[{Li et~al(2021)Li, Guo, Wang, and Zheng}]{Li2021}
Li D, Guo H, Wang Z, et~al (2021) {Unsupervised Fake News Detection Based on
  Autoencoder}. IEEE Access 9:29,356--29,365. \doi{10.1109ACCESS.2021.3058809}

\bibitem[{Liu and Wu(2020)}]{Liu2020}
Liu Y, Wu YFB (2020) {FNED A Deep Network for Fake News Early Detection on
  Social Media}. ACM Transactions on Information Systems 38(3):1--33.
  \doi{10.11453386253}

\bibitem[{Lundberg and Lee(2017)}]{Lundberg:2017}
Lundberg SM, Lee SI (2017) {A Unified Approach to Interpreting Model
  Predictions}. In: Proceedings of the International Conference on Neural
  Information Processing Systems. Curran Associates Inc., pp 4768--4777,
  \doi{10.5555/3295222.3295230}

\bibitem[{Mahajan et~al(2021)Mahajan, Shah, and Jafar}]{Mahajan:2021}
Mahajan A, Shah D, Jafar G (2021) {Explainable AI Approach Towards Toxic
  Comment Classification}. In: Proceedings of the Emerging Technologies in Data
  Mining and Information Security Conference. Springer, pp 849--858,
  \doi{10.1007/978-981-33-4367-2_81}

\bibitem[{Martens et~al(2018)Martens, Aguiar, Gomez, and
  Mueller-Langer}]{martens2018digital}
Martens B, Aguiar L, Gomez E, et~al (2018) {The Digital Transformation of News
  Media and the Rise of Disinformation and Fake News}. SSRN Electronic Journal
  pp 1--58. \doi{10.2139/ssrn.3164170}

\bibitem[{Mathew et~al(2021)Mathew, Amudha, and Sivakumari}]{Mathew2021}
Mathew A, Amudha P, Sivakumari S (2021) {Deep Learning Techniques: An
  Overview}, vol 1141. Springer, \doi{10.1007/978-981-15-3383-9_54}

\bibitem[{Mosallanezhad et~al(2022)Mosallanezhad, Karami, Shu, Mancenido, and
  Liu}]{Mosallanezhad2022}
Mosallanezhad A, Karami M, Shu K, et~al (2022) {Domain Adaptive Fake News
  Detection via Reinforcement Learning}. In: Proceedings of the ACM Web
  Conference. Association for Computing Machinery, pp 3632--3640,
  \doi{10.1145/3485447.3512258}

\bibitem[{Nasir et~al(2021)Nasir, Khan, and Varlamis}]{Nasir2021}
Nasir JA, Khan OS, Varlamis I (2021) {Fake news detection A hybrid CNN-RNN
  based deep learning approach}. International Journal of Information
  Management Data Insights 1(1):100,007--100,019.
  \doi{10.1016j.jjimei.2020.100007}

\bibitem[{Nikiforos et~al(2020)Nikiforos, Vergis, Stylidou, Augoustis,
  Kermanidis, and Maragoudakis}]{Nikiforos2020}
Nikiforos MN, Vergis S, Stylidou A, et~al (2020) {Fake News Detection Regarding
  the Hong Kong Events from Tweets}, vol 585 IFIP. Springer,
  \doi{10.1007/978-3-030-49190-1_16}

\bibitem[{Pham et~al(2017)Pham, Dang, Dinh, Hoang, Nguyen, and Liew}]{Pham2017}
Pham XC, Dang MT, Dinh SV, et~al (2017) {Learning from Data Stream Based on
  Random Projection and Hoeffding Tree Classifier}. In: Proceedings of the
  International Conference on Digital Image Computing: Techniques and
  Applications, vol 2017-Decem. IEEE, pp 1--8, \doi{10.1109/DICTA.2017.8227456}

\bibitem[{Puraivan et~al(2021)Puraivan, Godoy, Riquelme, and
  Salas}]{Puraivan2021}
Puraivan E, Godoy E, Riquelme F, et~al (2021) {Fake news detection on Twitter
  using a data mining framework based on explainable machine learning
  techniques}. In: Proceedings of the International Conference of Pattern
  Recognition Systems. Institution of Engineering and Technology, pp 157--162,
  \doi{10.1049/icp.2021.1450}

\bibitem[{Ribeiro et~al(2016)Ribeiro, Singh, and Guestrin}]{Ribeiro:2016}
Ribeiro MT, Singh S, Guestrin C (2016) {Why Should I Trust You Explaining the
  predictions of any classifier}. In: Proceedings of the ACM SIGKDD
  International Conference on Knowledge Discovery and Data Mining. Association
  for Computing Machinery, pp 1135--1144, \doi{10.11452939672.2939778}

\bibitem[{Shu(2022)}]{Shu:2022}
Shu K (2022) {Combating disinformation on social media: A computational
  perspective}. BenchCouncil Transactions on Benchmarks, Standards and
  Evaluations 2(1):100,035--100,040. \doi{10.1016/j.tbench.2022.100035}

\bibitem[{Shu et~al(2017)Shu, Sliva, Wang, Tang, and Liu}]{Shu2017}
Shu K, Sliva A, Wang S, et~al (2017) {Fake News Detection on Social Media A
  Data Mining Perspective}. SIGKDD Explorations Newsletter 19(1):22--36.
  \doi{10.11453137597.3137600}

\bibitem[{Shu et~al(2019{\natexlab{a}})Shu, Cui, Wang, Lee, and Liu}]{Shu:2019}
Shu K, Cui L, Wang S, et~al (2019{\natexlab{a}}) {dEFEND: Explainable Fake News
  Detection}. In: Proceedings of the ACM SIGKDD International Conference on
  Knowledge Discovery \& Data Mining. Association for Computational
  Linguistics, pp 395--405, \doi{10.1145/3292500.3330935}

\bibitem[{Shu et~al(2019{\natexlab{b}})Shu, Wang, and Liu}]{Shu2019Beyond}
Shu K, Wang S, Liu H (2019{\natexlab{b}}) {Beyond News Contents The Role of
  Social Context for Fake News Detection}. In: Proceedings of the ACM
  International Conference on Web Search and Data Mining. Association for
  Computing Machinery, pp 312--320, \doi{10.11453289600.3290994}

\bibitem[{Silva et~al(2021{\natexlab{a}})Silva, Han, Luo, Karunasekera, and
  Leckie}]{Silva:2021}
Silva A, Han Y, Luo L, et~al (2021{\natexlab{a}}) {Propagation2Vec: Embedding
  partial propagation networks for explainable fake news early detection}.
  Information Processing \& Management 58(5):102,618--102,634.
  \doi{10.1016/j.ipm.2021.102618}

\bibitem[{Silva et~al(2021{\natexlab{b}})Silva, Fontes, and
  J{\'{u}}nior}]{Caio2021}
Silva CVM, Fontes RS, J{\'{u}}nior MC (2021{\natexlab{b}}) {Intelligent Fake
  News Detection A Systematic Mapping}. Journal of Applied Security Research
  16(2):168--189. \doi{10.108019361610.2020.1761224}

\bibitem[{Silva et~al(2020)Silva, Santos, Almeida, and Pardo}]{Silva2020}
Silva RM, Santos RL, Almeida TA, et~al (2020) {Towards automatically filtering
  fake news in Portuguese}. Expert Systems with Applications
  146:113,199--113,212. \doi{10.1016/j.eswa.2020.113199}

\bibitem[{Sinaga and Yang(2020)}]{Sinaga2020}
Sinaga KP, Yang MS (2020) {Unsupervised K-Means Clustering Algorithm}. IEEE
  Access 8:80,716--80,727. \doi{10.1109ACCESS.2020.2988796}

\bibitem[{{\v{S}}krlj et~al(2021){\v{S}}krlj, Martinc, Lavra{\v{c}}, and
  Pollak}]{Skrlj2021}
{\v{S}}krlj B, Martinc M, Lavra{\v{c}} N, et~al (2021) {autoBOT: evolving
  neuro-symbolic representations for explainable low resource text
  classification}. Machine Learning 110(5):989--1028.
  \doi{10.1007/s10994-021-05968-x}

\bibitem[{Song et~al(2021)Song, Shu, and Wu}]{Song2021}
Song C, Shu K, Wu B (2021) {Temporally evolving graph neural network for fake
  news detection}. Information Processing \& Management 58(6):102,712--102,729.
  \doi{10.1016j.ipm.2021.102712}

\bibitem[{Sutton and Barto(2018)}]{sutton2018reinforcement}
Sutton RS, Barto AG (2018) {Reinforcement learning: An introduction}. MIT Press

\bibitem[{Tandoc(2019)}]{tandoc2019facts}
Tandoc EC (2019) {The facts of fake news: A research review}. Sociology Compass
  13(9):12,724--12,732. \doi{10.1111/soc4.12724}

\bibitem[{Vicario et~al(2019)Vicario, Quattrociocchi, Scala, and
  Zollo}]{Vicario2019}
Vicario MD, Quattrociocchi W, Scala A, et~al (2019) {Polarization and Fake News
  Early Warning of Potential Misinformation Targets}. ACM Transactions on the
  Web 13(2):1--22. \doi{10.11453316809}

\bibitem[{Vouros et~al(2021)Vouros, Langdell, Croucher, and
  Vasilaki}]{Vouros2021}
Vouros A, Langdell S, Croucher M, et~al (2021) {An empirical comparison between
  stochastic and deterministic centroid initialisation for K-means variations}.
  Machine Learning 110(8):1975--2003. \doi{10.1007/s10994-021-06021-7}

\bibitem[{Wang et~al(2020)Wang, Yang, Ma, Xu, Zhong, Deng, and Gao}]{Wang2020}
Wang Y, Yang W, Ma F, et~al (2020) {Weak Supervision for Fake News Detection
  via Reinforcement Learning}. Proceedings of the AAAI Conference on Artificial
  Intelligence 34:516--523. \doi{10.1609/aaai.v34i01.5389}

\bibitem[{Xiao et~al(2020)Xiao, Li, Qiang, Li, Xiao, and Liu}]{Xiao2020}
Xiao Y, Li W, Qiang S, et~al (2020) {A Rumor \& Anti-rumor Propagation Model
  Based on Data Enhancement and Evolutionary Game}. IEEE Transactions on
  Emerging Topics in Computing 10(2):690--703. \doi{10.1109/TETC.2020.3034188}

\bibitem[{Xue et~al(2021)Xue, Zhu, and Wang}]{Xue2021}
Xue Q, Zhu Y, Wang J (2021) {Joint Distribution Estimation and Na{\"{i}}ve
  Bayes Classification Under Local Differential Privacy}. IEEE Transactions on
  Emerging Topics in Computing 9(4):2053--2063. \doi{10.1109/TETC.2019.2959581}

\bibitem[{Ying et~al(2021)Ying, Yu, Wang, Ji, and Qian}]{Ying2021}
Ying L, Yu H, Wang J, et~al (2021) {Multi-Level Multi-Modal Cross-Attention
  Network for Fake News Detection}. IEEE Access 9:132,363--132,373.
  \doi{10.1109/ACCESS.2021.3114093}

\bibitem[{Zhao et~al(2020)Zhao, Zhao, Sano, Levy, Takayasu, Takayasu, Li, Wu,
  and Havlin}]{Zhao2020}
Zhao Z, Zhao J, Sano Y, et~al (2020) {Fake news propagates differently from
  real news even at early stages of spreading}. EPJ Data Science 9(1):7--20.
  \doi{10.1140/epjds/s13688-020-00224-z}

\bibitem[{Zhou et~al(2020)Zhou, Jain, Phoha, and Zafarani}]{Zhou2020Theory}
Zhou X, Jain A, Phoha VV, et~al (2020) {Fake News Early Detection A
  Theory-driven Model}. Digital Threats Research and Practice 1(2):1--25.
  \doi{10.1145/3377478}

\bibitem[{Zubiaga et~al(2017)Zubiaga, Liakata, and Procter}]{Zubiaga2017}
Zubiaga A, Liakata M, Procter R (2017) {Exploiting Context for Rumour Detection
  in Social Media}. In: Lecture Notes in Computer Science (including subseries
  Lecture Notes in Artificial Intelligence and Lecture Notes in
  Bioinformatics), vol 10539 LNCS. Springer, p 109--123,
  \doi{10.1007/978-3-319-67217-5_8}

\end{thebibliography}

\end{document}